\documentclass{article}
\usepackage{ijcai24}

\usepackage{times}
\usepackage{soul}
\usepackage{url}
\usepackage[hidelinks]{hyperref}
\usepackage[utf8]{inputenc}
\usepackage[small]{caption}
\usepackage{graphicx}
\usepackage{subcaption}
\usepackage{amsmath}
\usepackage{amsthm}
\usepackage{amssymb}
\usepackage{booktabs}
\usepackage{algorithm}
\usepackage{algorithmic}
\usepackage[switch]{lineno}
\usepackage{multirow}

\title{CIER: A Novel Experience Replay Approach with Causal Inference in Deep Reinforcement Learning}

\author{
$^1$
Jingwen Wang\and
Dehui Du$^1$\and
Yida Li$^1$\and
Yiyang Li$^1$
\And
Yikang Chen$^4$\\
\affiliations
$^1$East China Normal University\\
\emails
51265902037@stu.ecnu.edu.cn,
dhdu@sei.ecnu.edu.cn,
\{10215101567, 51215902128, 51265902150\}@stu.ecnu.edu.cn
}

\begin{document}

\maketitle

\begin{abstract}
In the training process of Deep Reinforcement Learning (DRL), agents require repetitive interactions with the environment. With an increase in training volume and model complexity, it is still a challenging problem to enhance data utilization and explainability of DRL training. This paper addresses these challenges by focusing on the temporal correlations within the time dimension of time series. We propose a novel approach to segment multivariate time series into meaningful subsequences and represent the time series based on these subsequences. Furthermore, the subsequences are employed for causal inference to identify fundamental causal factors that significantly impact training outcomes. We design a module to provide feedback on the causality during DRL training. Several experiments demonstrate the feasibility of our approach in common environments, confirming its ability to enhance the effectiveness of DRL training and impart a certain level of explainability to the training process. Additionally, we extended our approach with priority experience replay algorithm, and experimental results demonstrate the continued effectiveness of our approach.
\end{abstract}


\section{Introduction}
\textit{Reinforcement Learning} (RL) enables agents to iteratively interact with the environment, acquiring an optimal policy to maximize the cumulative expected reward. Due to the intricacies of real-world environmental states, RL simplifies the environment's state transition model through a Markov decision process. Consequently, most RL models adhere to the Independent and Identically Distributed (i.i.d.) assumption. To satisfy this assumption, experience replay \cite{lin1992self} is extensively employed in offline policy algorithms to store past experiences in a replay buffer. During training, random samples are drawn from this replay buffer to break the temporal correlation between consecutive experiences, improving the stability of learning. Thereby facilitating more efficient updates of model parameters. In light of this, deep Q networks \cite{mnih2013playing} integrate Q-learning, stochastic minibatch updates, and experience replay mechanisms, presenting a \textit{Deep Reinforcement Learning} (DRL) model applicable for the first time to high-dimensional data.


However, with the development of DRL, two challenges have emerged, namely, \textit{insufficient data efficiency}, and \textit{inadequate explainability} \cite{dulac2019challenges,survey2023crl}. In more complex application domains, such as robotics control and autonomous driving, sample efficiency has become a significant hurdle due to sparse rewards and the imbalanced distribution of observations in a large state space \cite{kiran2021deep}. Currently, various approaches are being explored to enhance sample efficiency in DRL, including reward shaping \cite{ng1999policy}, imitation learning \cite{christiano2017deep}, transfer learning \cite{vecerik2017leveraging}, meta-learning \cite{wang2016learning}, etc.


Meanwhile, the explainability of DRL models also significantly constrains their applicability in handling complex tasks. As far as we are aware, the research approaches in eXplainable Artificial Intelligence (XAI) can be categorized into transparent approaches and post-hoc explainability \cite{arrieta2020explainable,heuillet2021explainability}. In the context of DRL, standard algorithms are often non-transparent. Consequently, several studies have proposed explainable approaches tailored to specific domains. For instance, representation learning approaches focused on states, actions, and policies have been introduced \cite{raffin2019decoupling}. Simultaneous learning of policies and explanations has been explored \cite{juozapaitis2019explainable}, as well as multi-objective learning approaches \cite{beyret2019dot,cideron2019self} based on hindsight experience replay \cite{andrychowicz2017hindsight}. Additionally, there are approaches using Shapley value for global reward allocation explanations in multi-agent systems \cite{wang2020shapley}. In post-hoc explainability, some studies treat DRL algorithms as black boxes and propose explanation approaches, such as saliency maps in image data \cite{selvaraju2017grad}. 


According to the latest research advancements, causal inference \cite{pearl2018book} represents a promising research direction distinct from traditional statistics. Different from conventional statistical approaches, causal reasoning is closely aligned with human cognitive psychology, making it more explainable and in harmony with our thought processes. This characteristic lends itself to a stronger explanatory power, aligning well with principles from cognitive psychology in human cognition \cite{kuang2020causal,young2016unifying}. To the best of our knowledge, there is scant research dedicated to explaining the training process of DRL, particularly the experience replay process, through the lens of causal inference. 


In this paper, we aim to propose an approach that combines time series analysis and causal inference to enhance sample efficiency in DRL and provide explainability for the training process. We posit that there are numerous repetitive ``patterns" in the training process of DRL, and these patterns may exhibit temporal correlations. Therefore, we segment the interaction data between the environment and the agent into explainable patterns. Based on these patterns, we represent historical training trajectories. To uncover hidden temporal correlations in the data, we consider these ``meaningful patterns" as temporal factors and employ causal reasoning approaches to construct a causal graph for these factors. To integrate the discovered causality into the DRL training process, we introduce a novel experience replay architecture. This architecture allows the consideration of causality between policy goals and temporal factors during the experience replay process. Additionally, it enhances \textit{Prioritized experience replay} (PER) \cite{schaul2015prioritized}. We conducted experiments in common DRL environments, demonstrating the effectiveness and scalability of our approach.


In a nutshell, the contributions of this work are as follows:


\begin{itemize}
\item We introduce a representation approach for multivariate time series that focuses on capturing internal correlations within the time series. This approach segments time series into explainable patterns, using these patterns as fundamental units to represent the original data.

\item In the field of DRL, we introduce a novel experience replay approach, referred to as \textit{Causal Inference Experience Replay} (CIER). This approach employs causal inference techniques to analyze the causality within historical experience data in DRL, utilizing causal effects as a metric for prioritizing the replay of experiences.

\item We show that our approach leads to performance improvements in several state-of-the-art DRL models. Furthermore, our approach demonstrates a certain degree of scalability. We enhance priority experience replay with CIER and validate its effectiveness through experiments.

\end{itemize}


\section{Related Work}

\subsection{Multivariate Time Series Representation}

During the DRL training process, time series can become highly intricate. To enhance analytical efficiency, we opt for dimensionality reduction, which serves as the motivation for representation. Currently, three commonly used representation approaches include piecewise approximation, data representation through identifying important points, and symbolic representation \cite{wilson2017data}. The Piecewise approximation algorithm segments the time series into non-overlapping sub-intervals and fits local models within each interval \cite{keogh1997probabilistic,lee2003dimensionality,latecki2005partial}. Preserving globally significant points aids in filtering information worthy of retention \cite{fu2001pattern,5234775}, while symbolic representation involves dividing the data into different regions and symbolically representing each region \cite{lin2003symbolic,lin2007experiencing}. However, these approaches often suffer from a lack of explainability. Therefore, we aim to uncover latent information in the data through data mining approaches. In the domain of pattern recognition, many approaches based on the matrix profile have demonstrated excellent performance \cite{yeh2016matrix,yeh2017matrix}. In clustering approaches, David Hallac proposed the \textit{Toeplitz Inverse Covariance-based Clustering} (TICC) approach \cite{hallac2017toeplitz}, utilizing Markov random fields to define each cluster on multivariate time series, showcasing remarkable explainability. 


\subsection{Causal Discovery in Time Series}
Identifying causality is a fundamental issue in time series data mining, however, due to the high dimensionality and long sequence issues that may exist in multivariate time series, it is challenging to discover causality in multivariate time series. 

A classic technique for determining the relationship between time series dimensions is the Granger causality test, which was proposed by Clive Granger \cite{granger1969investigating}. Another mainstream causal discovery algorithm is based on the concept of conditional independence \cite{pearl2000models}, such as the Peter-Clark algorithm \cite{spirtes1991algorithm} and \textit{Fast Causal Inference} (FCI) algorithm \cite{spirtes2000causation}. Building upon previous work, the \textit{Greedy Fast Causal Inference} (GFCI) algorithm is introduced to address the effect of unmeasured confounders. Independent component analysis \cite{hyvarinen2000independent} has also been extended to find causality in multivariate time series. However, the majority of existing work is confined to seeking causality between features while overlooking temporal internal correlations.



\subsection{Experience Replay}
The formal study of experience replay was initiated by Lin \textit{et al.} \cite{lin1992self}, then widely used in various DRL models \cite{mnih2013playing,mnih2015human,silver2014deterministic,mnih2016asynchronous}. A related issue, which is raised by the uniform random sampling approach of the classic experience replay algorithm, is that the distinctions in relevance across experiences are masked, and the update effect can be improved \cite{schaul2015prioritized}.

Fortunately, several studies have enhanced the experience replay algorithm to solve this flaw. Schaul \textit{et al.} proposed a priority experience replay algorithm based on \textit{Temporal Difference} (TD) error \cite{schaul2015prioritized}. Andrychowicz \textit{et al.} presented hindsight experience replay \cite{andrychowicz2017hindsight}, which enables sampling-effective learning from sparse and binary rewards. Zhang \textit{et al.} noted the significant impact of the replay buffer size on results and proposed combined experience replay to mitigate the effects of a large replay buffer \cite{zhang2017deeper}. Building on this, Fedus \textit{et al.} demonstrated that increasing replay capacity and reducing the age of the oldest policy can enhance agent performance \cite{fedus2020revisiting}.

Our approach does not rely on traditional dimension correlations. Instead, we analyze meaningful time subsequences in the temporal dimension and enhance the experience replay method from a causal inference perspective. Noting that the majority of DRL models that employ the experience replay approach can benefit from our research.



\section{Preliminaries}
A standard DRL formalism can be represented by a tuple $(S,A,P_t,R,\gamma)$, which contains a set of states $S$, a set of actions $A$, the transition probabilities $P_t: S \times A \to S$, the reward function $R : S \times A \times S \to R$ and the discount factor $\gamma \in (0,1)$. And the experience in DRL can be defined as the transition tuple $(s_t,a_t,r_t,s_{t+1})$. 
In order to mine time internal relationships from prior experience, the trajectory of each training of the agent can be extracted as an action time series. Consider a multivariate action time series of $T_A = \left[a_1,a_2,\dots,a_n\right] \in \mathbb{R}^n$, 
where $n$ is the length of $T_A$ and $a_i\in \mathbb{R}$ is the $i$-th multivariate action vector of $T_A$. Formally, a subsequence, which is a continuous subset of $T_A$, can be defined as $T_{i,j} = [a_i,a_{i+1},…,a_{j-1},a_{j}]$, where $0\leq i<j\leq n$.

Causal discovery algorithms have the capability to identify not only correlations but also causality within observation. FCI is a constraint-based algorithm that falls within this category. FCI can discover unobserved confounding factors. It starts with a fully undirected graph corresponding to the space of observable structures, and based on d-separation criteria, it prunes edges to eventually generate a \textit{Partial Ancestral Graph} (PAG).


In the presence of unobserved variables, PAG represents the uncertainty of the joint probability distribution of observation. In brief, PAG models four types of relationships:


\begin{itemize}
    \item $A \to B $: $A$ is a cause of $B$. 
    \item $A \leftrightarrow B$ : there is no causality between $A$ and $B$ and there must be an unobserved confounding variable that is a parent node of both $A$ and $B$.
    \item $A \circ\rightarrow B$: on the basis of $A \leftrightarrow B$, it is possible that $ A $ is the cause of $B$. 
    \item $A \circ-\circ B$: on the basis of $A \circ\to B$, it is possible that $B$ is the cause of $A$.
\end{itemize}

In the study of causality,\textit{ Average Treatment Effect} (ATE) is a commonly used causal estimand and it can be defined as follows: $ATE=\mathbb{E}[Y_{\alpha=1}-Y_{\alpha=0}]$, where $\alpha = 1$ and $\alpha = 0$ represent the presence or absence of treatment respectively.


GFCI is an extension of FCI. In the first stage, it searches the empty observational structure space based on greedy equivalent search, maximizing the Bayesian score by adding edges and selecting the PAG with the highest score as the input for the second stage of FCI. It addresses the issue of high sample requirements in causal discovery and demonstrates excellent performance.


\section{Methodology Design}

In this section, we present the algorithmic details of CIER approach. We extract historical time sequences from the DRL training process, represent the time sequences based on clustering approaches, mine internal relationships within the time sequences, and employ causal discovery approaches to identify subsequences that lead to changes in rewards. This information is then fed back into the experience extraction process to enhance the efficiency of experience replay during the training process.


\subsection{Time Series Representation}
Our work aims to enhance the efficiency and explainability of the DRL training process. Accordingly, we abstract DRL historical actions into multivariate time series. For each training iteration, we arrange all action frames chronologically to construct the action time series $T_A$. With the increase in training iterations and model complexity, the lengths of sequences for each training iteration may exhibit significant variations. Presently, numerous time series analysis algorithms impose equal-length requirements on time series lengths or employ approaches such as truncation, interpolation, and padding. Such practices undoubtedly introduce direct interference with the information encapsulated in the time series.


Therefore, we propose a representation approach for time series based on \textit{Time Series Causal Factors} (TSCFs). We make an extension to subsequences $T_{i,j}$, that is, if we can partition a time series into meaningful subsequences, and these subsequences can characterize each time series in the collection of time series, then we define the aforementioned subsequences as TSCF, which has the following properties.


\paragraph{Time Series Causal Factor (TSCF)}
For each $T$ in the collection of time series, a TSCF $f_i$ is a meaningful subsequence of $T$, where $T = \bigcup f_i$, and $f_i\subset T_{i,j} $.

For each TSCF, it can be abstracted and converted into a binary causal variable that may have a causal effect with each other in the structural causal model.

\begin{figure}[h]
	\includegraphics[trim= 1cm 0 0 0, clip, width=0.46\textwidth]{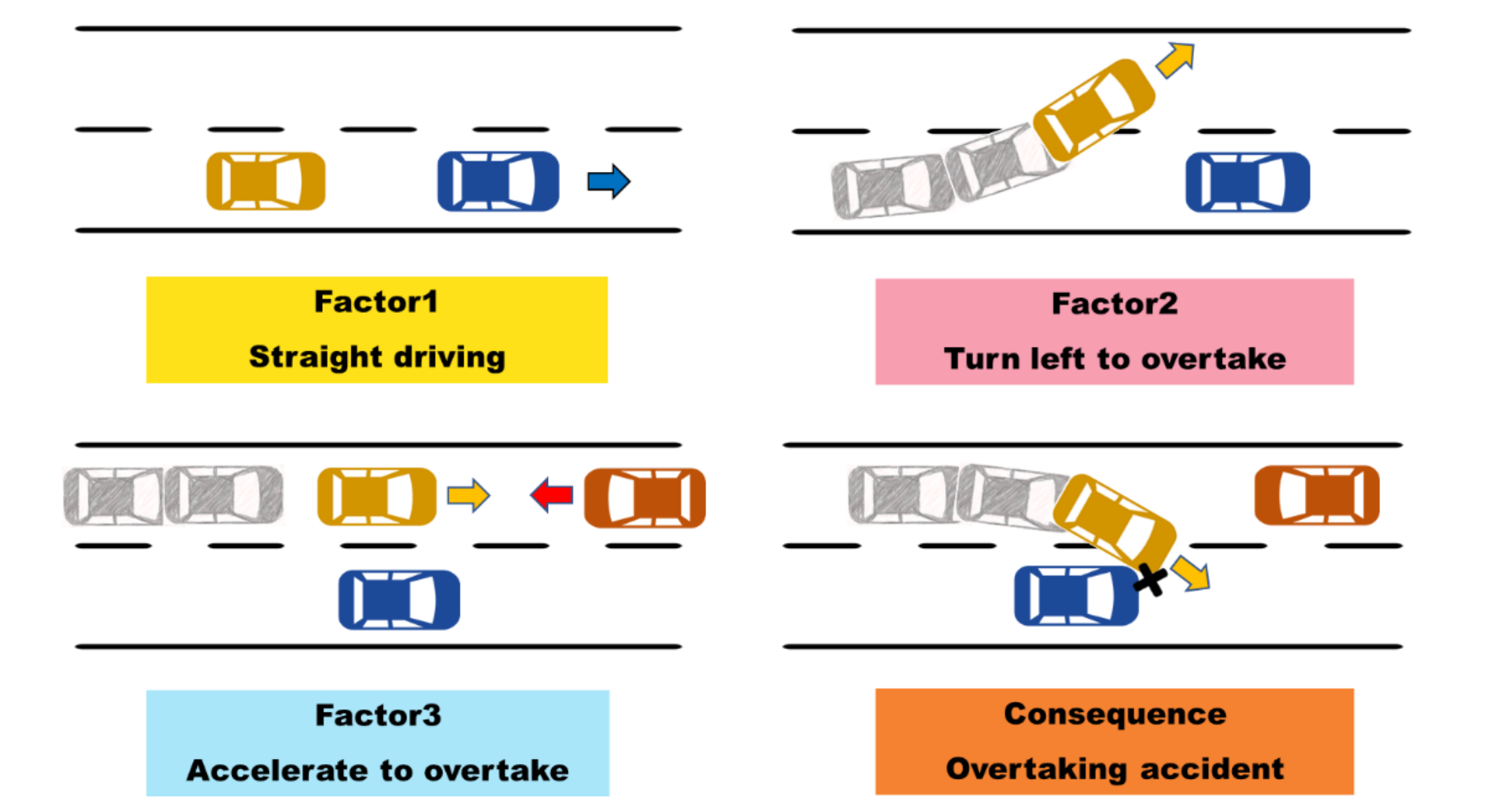}
	\caption{
A critical overtaking scenario. The entire scenario can be divided into four TSCFs based on the temporal order of actions. \textit{Factor 1} represents following the vehicle straight, \textit{Factor 2} represents the driver's intention to overtake by turning left into the oncoming lane, \textit{Factor 3} represents the driver's intention to accelerate and complete the overtaking action before a collision occurs, and the final TSCF (Consequence) represents a misjudgment by the driver in terms of distance and speed, resulting in a collision.}
	\label{fig:overtake}
\end{figure}

In order to demonstrate the role of TSCF, we conducted a critical scenario study in the field of autonomous driving to illustrate the importance of the correlations among TSCFs, as depicted in Figure \ref{fig:overtake}.


If we intend to summarize the errors in Figure \ref{fig:overtake} and avoid similar collisions in future scenarios, it is evident that we would not independently consider dimensions such as the direction or speed of the vehicle in influencing the consequence. Instead, we might consider actions like deceleration and avoidance in \textit{Factor 3} or trace back to more fundamental reasons: abandoning the overtaking action at \textit{Factor 2}. In summary, we should treat time series subsequences as fundamental units for investigation, where a meaningful subsequence can enhance the explainability of research findings.

\begin{figure}[h]
	\centering
	\includegraphics[trim= 0.3cm 0.2cm 0cm 0cm, clip, width=0.33\textwidth]{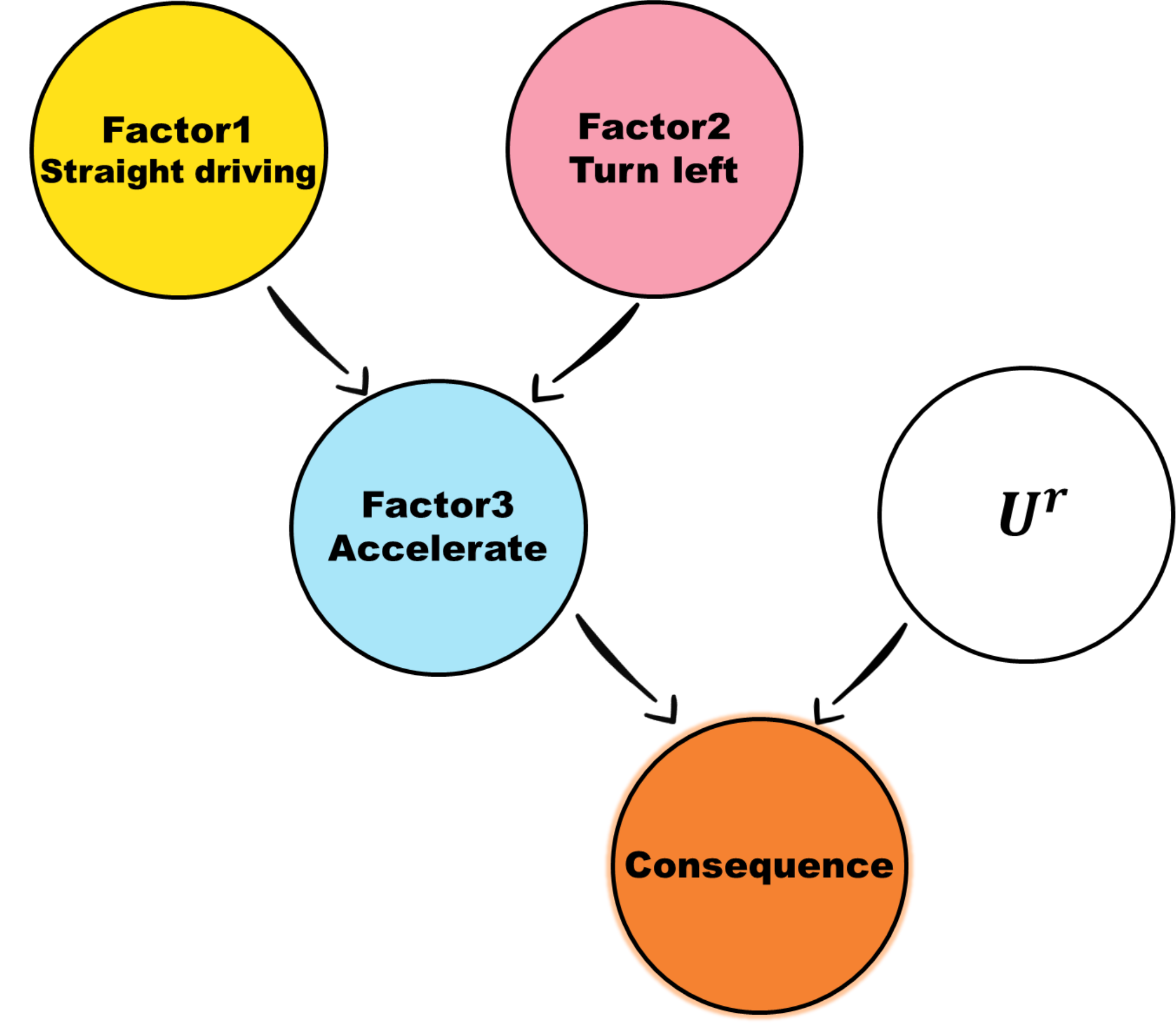}
	\caption{Causal graph of overtaking scenario based on TSCFs.}
	\label{fig:causal_graph}
\end{figure}

However, considering the impact of TSCF on the target in isolation is limited. As depicted in Figure \ref{fig:causal_graph}, we constructed a causal graph based on TSCFs and assumed that factors in the environment, apart from the ego vehicle's behavior, remain constant. These factors are referred to as exogenous variables, labeled as $U_r$. Therefore, \textit{Factor 3} and \textit{Factor 2} are considered the direct and indirect causes of the outcome, respectively.


If we modify the actions in \textit{Factor 3}, either accelerating or decelerating decisions have a probability of avoiding accidents caused by overtaking. However, if we modify the action in \textit{Factor 2} and assume the left-turn action is unsuccessful, \textit{Factor 3} will not occur, and the accident outcome caused by overtaking will certainly not happen. From the perspective of a DRL agent, paying more attention to \textit{Factor 2} would bring more benefits to the training process, reducing rewards and penalties resulting from accidents. On the other hand, excessive emphasis can lead to biases in training strategies, a point we will address later.


We identified TSCFs in DRL action time series based on TICC \cite{hallac2017toeplitz}. The objective of TICC is to cluster the observation values $T_A$ into $K$ clusters. To consider the correlations between variables during clustering, each cluster is represented by a \textit{Precision Matrix} (PM) $\Theta_i \in R^{dw_i \times dw_i}$, where \(d\) is the feature dimension of $T_A$, and \(w_i\) represents the sequence length of the $i\text{-}th \; T_A$. A PM can be obtained by computing the inverse of the covariance matrix, and it can represent the conditional independence structure between different clusters \cite{koller2009probabilistic}. 
Formally, $\theta_{i,j} = 0 \Leftrightarrow \text{COV}(X_i, X_j) = 0 \Leftrightarrow X_i \bot X_j | X_{V \backslash \{i,j\}}$, where \(X_i\) represents the $i\text{-}th$ cluster in the PM and \(V\) represents the set of all clusters..


Due to the simultaneous need for global solutions for both assigning clusters and updating parameters for PM variables, expectation maximization is employed to iteratively perform clustering tasks and update cluster parameters. Upon convergence of the optimization task, the observation sequence set \(T_A\) is partitioned into \(K\) subsequences. Each subsequence is associated with a \(\Theta\) that describes the conditional independence relationships between nodes. Similarly, the local potential function of a Markov random field represents the relationships between a node and its adjacent nodes, and the global potential function is the product of all local potential functions. We consider \(\Theta\) as part of a graphical model representing dependencies between variables.



\begin{figure}[h]
	\centering
	\includegraphics[trim= 0cm 0cm 0cm 0cm, clip, width=0.48\textwidth]{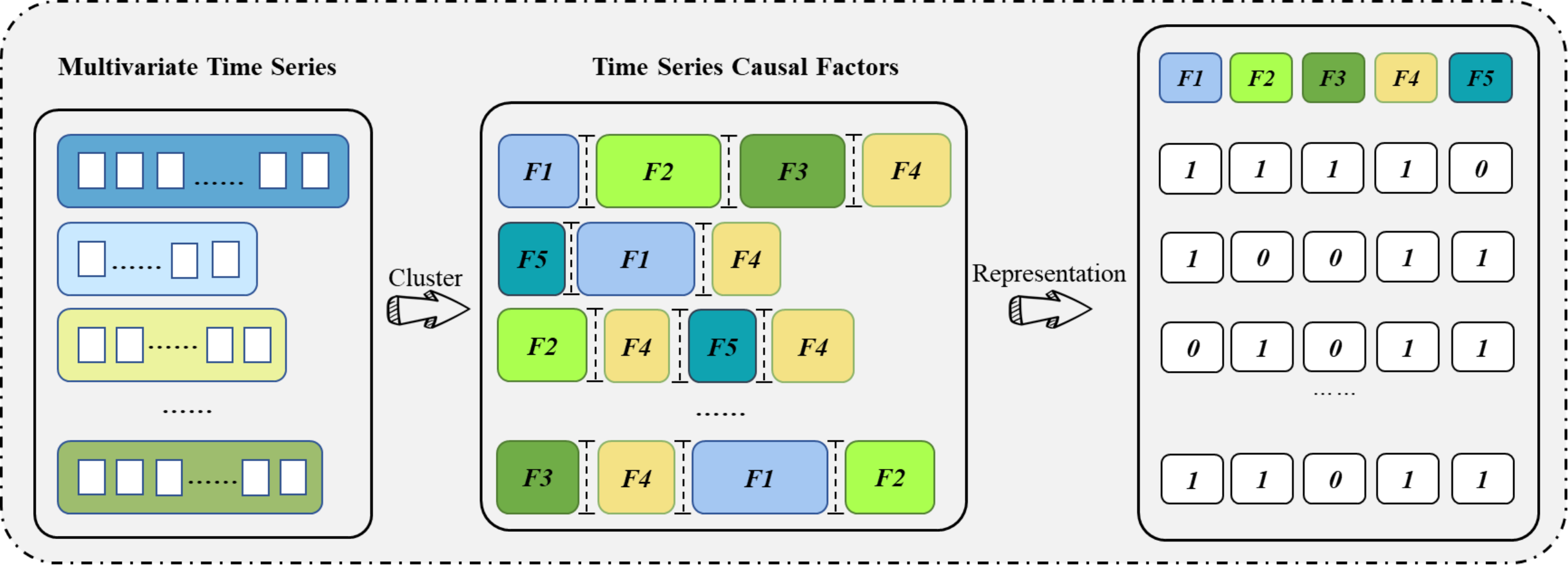}
	\caption{An illustration of the representation of multivariate time series. Unequal-length multivariate time series samples are transformed into equal-length sequences through the representation method of TSCF.}
	\label{fig:representation}
\end{figure}

After clustering individual multivariate time series, our goal is to merge clusters that exhibit high similarity among samples. Therefore, we apply the K-Means algorithm to perform a second round of clustering on all clusters based on the dynamic time warping metric and use TSCFs to represent $K'$ clustering results, that is, use a TSCF $f_i$ to represent a class of similar subsequence clusters. 
Furthermore, to simplify the representation, we use a binary variable $U_k \in \{0,1\}$ to indicate whether $T_{i,j}$ exists in this time series, where $T_{i,j}^k \subset f_k , k\in K'$. As shown in Figure \ref{fig:representation}, a multivariate time series can be expressed as a combination of $U_k$, formally, $T_A = \bigcup U_k$ , where $1\leq k \leq K'$. Due to the varying lengths of each time series, we dynamically adjust the value of \(K\) based on the time steps of the current \(T_A\), and set the value of \(K'\) to be the average of \(K\).


\subsection{Causality within Time Series}
In order to identify commonalities among the data, our approach diverges from traditional approaches by focusing on the temporal correlations within time series. We consider TSCF as the treatment, the reward obtain from each training iteration as the outcome variable, and construct a PAG based on the GFCI algorithm. In the process of causal discovery, the Bayesian information criterion is employed to assess the fitting quality and complexity score of causal graph models. Within the PAG, apart from ``$\to$" denoting deterministic causality, uncertainty exists for the remaining cases. 


To optimize the PAG, we handle the three remaining scenarios differently. In cases where it is uncertain whether a causality exists, we assertively assume the presence of the causality. While this might be an erroneous approach in other causal inference applications, our specific context aims to identify all potential sets of factors contributing to reward changes, rather than precisely determining the ``most appropriate" set of reward causes. The latter is a subset of the sets we identify, as explained later. Hence, we ignore the A $\leftrightarrow$ B relationship, simultaneously asserting the direct causality A $o\to$ B. For the A$o-o$B relationship, we initially assume causality between A and B. We then calculate the probabilities of A appearing before B and B appearing before A in the observation. 


The underlying logic stems from Granger causality test principle: future events will not have causal effects on the present and past, only past events may affect the present and future. Consequently, we choose the treatment with a higher probability of occurring first as the causal variable and the other as the outcome variable. To some extent, this approach mitigates the temporal information loss issue arising from representing time series with binary variables. We term this approach as ``time correction". Since the adoption of the presence or absence of TSCF as a representation in the temporal sequence, certain chronological order might be overlooked. Hence, in this context, we apply adjustments to the PAG in the temporal dimension.


For the modified PAG graph, we select all paths that end at the result node and intervene on all variables on the path to calculate their causal strength with the result node. If there exists an indirect causal effect between variables and the result, then we traverse all edges between them and accumulate their causal strengths as the causal strength. Through the above processing, we have screened out TSCFs that may cause changes in reward and calculated their causal effects.


\subsection{Causal Inference Experience Replay}

For the ordinary DRL experience replay algorithm, each round of training samples is sampled as experience, that is, transition $(s_t,a_t,r_t,s_{t+1})$, and stored in the queue. In order to reuse past experience and break the correlation of experience, one or more rounds are randomly selected for training to calculate the temporal difference and update parameters in combination with the gradient descent algorithm.


Our approach shares conceptual similarities with PER \cite{schaul2015prioritized}, as both aim to replace uniform sampling with non-uniform sampling, guiding experience sampling based on the differences in transition importance. In PER, transitions are prioritized according to the disparity between the current state and the expected state, \textit{i.e.}, the absolute value of TD error. In contrast, our approach is based on causality. As illustrated in Figure \ref{fig:overtake}, We believe that identifying the agent's behaviors that lead to changes in the target and focusing on the associated experiences can be more beneficial for training.


Therefore, in the previous algorithm, we analyzed the intelligent agent behavior sequence according to the change of reward and selected the treatment related to reward according to the PAG diagram. For the treatment in the PAG diagram, which corresponds to a TSCF in the behavior sequence and may be related to several consecutive actions, we assign the same weight to transitions containing these $a_t$. For different TSCFs, we assign related experiences with weights that are positively correlated with ATE according to the causal strength between the corresponding treatment and reward $\widetilde{w} \propto ATE$.


In the causal discovery phase, we modified the initial PAG diagram to obtain the maximum set of possible causality. Replaying the treatment from the set, even if it is not the real cause, does not have a detrimental impact on the results. Instead, it merely diminishes the efficiency improvement and convergence speed brought about by replay, in comparison to the real cause. Therefore, this explains why we choose to retain the maximum cause set.


In order to adapt to the training process in DRL environments, progressing from simple to complex, we employ a curriculum learning strategy to update the weights of experiences. Our strategy involves injecting more causal relationship weights in the early stages and reducing some causal influence in the later stages. The process is controlled by a hyperparameter \(u\), and its calculation formula is as follows:

\begin{tiny}
    \begin{equation}
    \mu = \frac{1}{\epsilon_{m}} \eta\sqrt{\epsilon_{m}^2-\epsilon_{c}^2} \, ,
    \end{equation}
\end{tiny}

\noindent where $\epsilon_{m}$ is the maximum number of training epochs, $\epsilon_{c}$ is the current training epoch, and $\eta$ represents the proportional constant for the curriculum learning control parameter.


\begin{figure}[h]
	\centering
	\includegraphics[trim= 0cm 0cm 0cm 0cm, clip, width=0.48\textwidth]{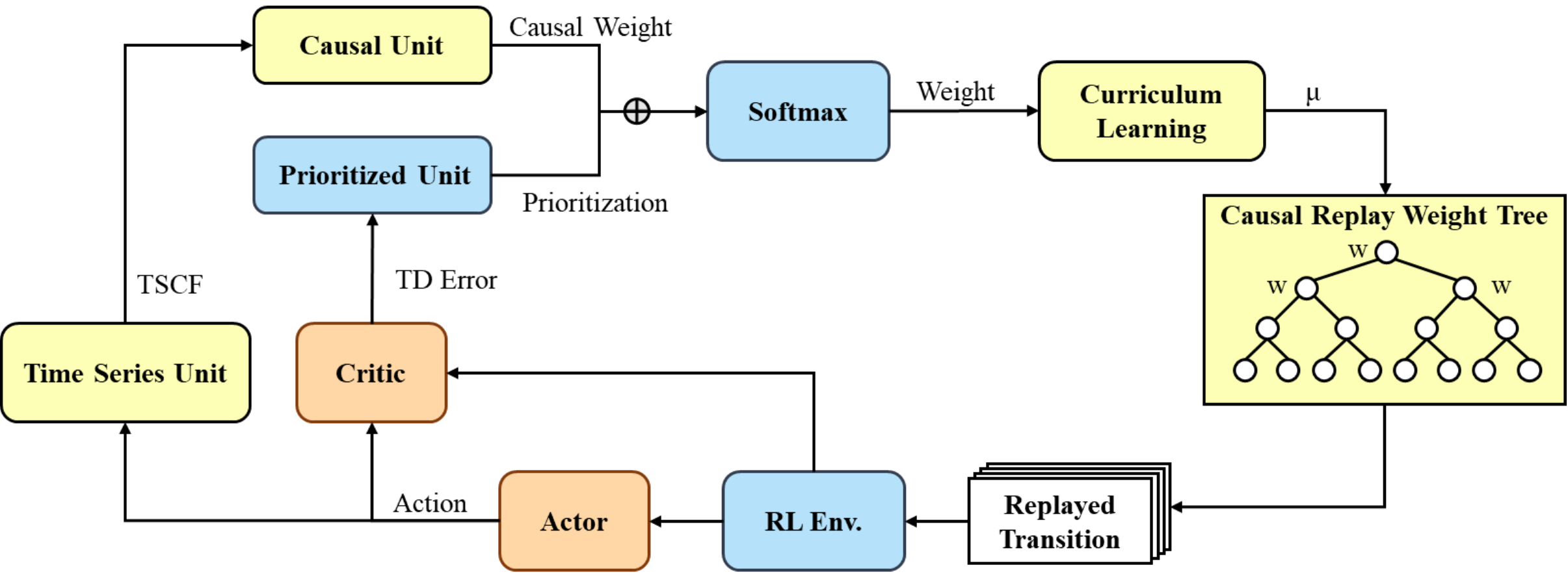}
	\caption{The architecture for CIER. It is based on the Actor-Critic algorithm and optimizes experience replay in the general DRL training process. During iterations, Time Series Unit represents the training historical data through TSCFs. The Causal Unit, based on the training target, conducts causal discovery on TSCF and maps causality to action sequences. The Prioritized Unit is optional and utilized for prioritizing experiences based on the TD Error calculated from the value function Estimation provided by the Critic.}
	\label{fig:cier}
\end{figure}

\paragraph{Causal Inference Experience Replay Model}

One of our contributions is that we have made improvements to the DRL experience replay process, as illustrated in Figure \ref{fig:cier}. The sum-tree structure \cite{schaul2015prioritized} (referred to as the Causal Replay Weight Tree) is employed to store the final experience weights, enhancing the efficiency of experience sampling. Experience weights are generated by internal causality and TD error. After linear weighting and controlled influence through curriculum learning, to prevent overfitting exploration caused by excessive focus on a particular TSCF, we blend the computed experience weight distribution with a uniform distribution. It ensures the integrity of the exploration space for experience replay. In Time Series Unit, the clustering counts \(K\) and \(K'\) are adaptively adjusted based on the length of each action time series. The algorithm is presented in Algorithm \ref{alg:algorithm}.


\begin{algorithm}[tb]
    \caption{Causal Inference Experience Replay Algorithm}
    \label{alg:algorithm}
    \textbf{Initialize}: The action time series $T_A$, the reward of each episode $\mathcal{R}$, experience replay buffer $\mathcal{B}$, time series causal factors $\mathcal{F}$, average treatment effect $ATE$, weights of transitions $\mathcal{W}$. 
    
    \begin{algorithmic}[1] 
        \WHILE{episode $\leq$ M}
        \WHILE{not Done}
        \STATE Construct $T_A$ from $\mathcal{B}$.
        \STATE $\mathcal{F} \gets K\text{-}Means(TICC(T_A, k))$.
        \STATE $PAG, ATE \gets GFCI(\mathcal{F},\mathcal{R})$.
        \STATE Modify the PAG with time correction.
        \STATE $\mathcal{W} \gets Curriculum \; learning(\mathcal{F},PAG, ATE)$.
        \IF {use PER}
        \STATE $\mathcal{W} \gets PER(TD \; error)$.
        \ENDIF
        \STATE Extract experiences and update the Critic networks and Actor networks.
        \ENDWHILE
        \ENDWHILE
    \end{algorithmic}
\end{algorithm}

\section{Experiment Evaluation}
Our approach can be applied to almost all environments in DRL where agents have multivariate continuous action spaces. In environments that necessitate discrete actions, we have also adjusted the action types in the environment. To evaluate the guiding role of CIER in the DRL training process and assess its scalability, we compared training performance with various environments.

\subsection{Experimental Setup}
\textbf{Datasets }  Our DRL environment is built upon the Highway-Env \cite{highway-env} \& Gym \cite{1606.01540} libraries. Highway-Env encompasses vehicle-based traffic flow and intricate highway scenarios, designed for testing and evaluating tasks related to traffic management, autonomous driving, and traffic flow control. It has been widely applied in the autonomous driving field. We conducted experiments in various environments within Highway-Env, including highway, intersection, and racetrack. The ego vehicle is tasked with avoiding collisions with other vehicles and completing specific tasks as soon as possible.
Consequently, the reward is formulated as a combination of speed and collision avoidance, as illustrated as follows, 

\begin{equation}
\label{equation:reward}
    R(s, a)=a \cdot \frac{v-v_{\min }}{v_{\max }-v_{\min }}-b \cdot\text { collision },
\end{equation}
where the collision penalty is set at -1, with $a$ and $b$ representing scaling coefficients. In Gym environments, we conducted experiments in the reacher and humanoid settings, both relying on the Mujoco engine and featuring continuous action spaces.


To more accurately uncover the causal relationships within temporal data, we increased the simulation frequency in the environment and made corresponding adjustments to the policy update frequency and the duration of action in our experiment. Apart from these modifications, no further changes were made to the standard environment.


\noindent \textbf{Baselines }
We employed two state-of-the-art baselines for training in various DRL environments: the \textit{Deep Deterministic Policy Gradient} algorithm (DDPG) \cite{lillicrap2015continuous} and the \textit{Twin Delayed Deep Deterministic Policy Gradient} algorithm (TD3) \cite{fujimoto2018addressing}. The Actor-Critic networks employed in our study consist of 3 to 4 fully connected layers, with a batch size of 256. To mitigate the potential impact of the replay buffer size on results, we fixed the size of the experience replay pool at 1,000,000. To enhance training efficiency and avoid recalculating TSCF in each iteration, we have designed a temporary buffer pool with a capacity of 100,000. Throughout the training process, the temporary buffer pool is synchronized with the experience replay pool to continuously update experiences. Only when the temporary buffer pool is filled will the causal analysis be re-run.


To evaluate the effectiveness of our approach, we initially compared the traditional random experience replay approach with our proposed causal inference experience replay approach. Furthermore, our experimental results indicate that CIER not only exhibits outstanding performance when used independently but also yields significant performance improvements when combined with PER algorithm.


\noindent\textbf{Metrics }
Based on prior research \cite{mnih2015human,schaul2015prioritized}, we employed widely used DRL training performance metrics, including \textit{Average Score} (AS) and \textit{Best Score} (BS). These metrics respectively reflect the real-time performance during the same training epochs and the achievable limit within the given training rounds. Additionally, we introduced two supplementary metrics, \textit{Step of Average Score} (SAS) and \textit{Average Cumulative Score} (ACS), which, to some extent, portray the convergence speed and overall performance throughout the entire task.

    \begin{itemize}
      \item $AS = \frac{1}{n} \sum_{i=1}^{n} \operatorname{Score}_i$
      \item $BS = \max_{1 \leq i \leq n} \operatorname{Score}_i$
      \item $SAS = \text{argmin}_{i} (\operatorname{Score}_i \geq AS) \, $
      \item $ACS =  \frac{1}{n} \sum_{i=1}^{n} \sum_{j=1}^{i} \operatorname{Score}_j$
    \end{itemize}

\subsection{Discussion}
To provide a more intuitive representation of the approach's effectiveness, our experiments focused on the results of the initial one thousand training rounds, and multiple trials were conducted to obtain the average values. In light of the current challenges in DRL, namely insufficient data efficiency and inadequate explainability, we conducted experiments to demonstrate the effectiveness of the CIER approach for addressing these issues.


\setlength{\aboverulesep}{1ex}
\setlength{\belowrulesep}{1ex}

\begin{table*}
    \centering
    \resizebox{\linewidth}{!}{
    \begin{tabular}{lrrrrrrrrrrrrrrrrrrrr}
        \toprule
        
        \multirow{2}{*}{\textbf{Models}} & \multicolumn{4}{c}{\textbf{Highway}} & \multicolumn{4}{c}{\textbf{Intersection}} &\multicolumn{4}{c}{\textbf{Racetrack}}  & \multicolumn{4}{c}{\textbf{Reacher}} & \multicolumn{4}{c}{\textbf{Humanoid}}\\
        \cmidrule(lr){2-5} \cmidrule(lr){6-9} \cmidrule(lr){10-13} \cmidrule(lr){14-17} \cmidrule(lr){18-21}
        & AS & BS & SAS & ACS & AS & BS & SAS & ACS & AS & BS & SAS & ACS & AS & BS & SAS & ACS & AS & BS & SAS & ACS \\
        \midrule
        DDPG & 24.92 & \textbf{28.87} & 159 & 11834.56 & 19.28 & \textbf{22.87} & 142 & 9187.02 & 19.76 & 26.9 & 516 & 9055.01 & -20.43	& -8.43	& 192 & -14679.13 & 281.53 & 429.75	& 377 & 115001.95 \\
        DDPG-CIER & \textbf{26.32} & 28.51 & \textbf{59} & \textbf{13136.85} & \textbf{19.63} & 22.68 & \textbf{194} & \textbf{9382.09} & \textbf{33.34} & \textbf{42.24} & \textbf{178} & \textbf{15392.79} & \textbf{-17.95} & \textbf{-7.73} & \textbf{190} & \textbf{-12706.25} & \textbf{324.21} & \textbf{454.33} & \textbf{338} & \textbf{133717.52} \\

        \midrule
        TD3 & 24.45 & 30.27 & 329 & 10658.83 & 15.64 & 18.69 & 400 & 7271.48 & 62.06 & 74.59 & 219 & 28153.3 & -13.6 & -11.18 & \textbf{110} & -7497.87 & 305.45 & 522.11 & \textbf{449} & 108124.2 \\
        TD3-CIER & \textbf{25.05} & \textbf{31.64} & \textbf{330} & \textbf{10869.48} & \textbf{16.59} & \textbf{19.26} & \textbf{357} & \textbf{7846.07} & \textbf{71.96} & \textbf{85.43} & \textbf{146} & \textbf{32660.13} & \textbf{-12.78} & \textbf{-9.3} & 184 & \textbf{-7309.49} & \textbf{342.91} & \textbf{688.52} & 507 & \textbf{119837.04} \\
        \midrule
        DDPG-PER & 25.81 & 28.58 & 123 & 12358.38 & 14.56 & 19.6 & 430 & 6527.1 & 31.02 & 36.8 & 297 & 14624.17 & -20.31 & -12.8 & 150 & -13120.67 & 241.9 & 407.34 & \textbf{344} & 91525.17 \\
        DDPG-CIPER & \textbf{26.89} & \textbf{29.46} & \textbf{136} & \textbf{13095.13} & \textbf{17.43} & \textbf{22.77} & \textbf{400} & \textbf{7660.87} & \textbf{38.3} & \textbf{50.23} & \textbf{295} & \textbf{16536.66} & \textbf{-17.86} & \textbf{-8.67} & \textbf{144} & \textbf{-12335.8} & \textbf{399.5} & \textbf{588.87} & 348 & \textbf{156935.79}\\
        \midrule
        TD3-PER &\textbf{26.84} & 29.45 & 326 & \textbf{12906.79} & 18.48 & \textbf{31} & 479 & 8825.26 & 57.08 & 66.65 & \textbf{229} & 26410.73 & -14.88 & -10.09 & 116 & -8371.55 & 299.13 & 531.94 & \textbf{414} & 115655.28\\
        TD3-CIPER & 24.27 & \textbf{30.06} & \textbf{314} & 10716.02 & \textbf{18.74} & 26.8 & \textbf{396} & \textbf{9132.75} & \textbf{64.53} & \textbf{78.61} & 268 & \textbf{29448.98} & \textbf{-12.18} & \textbf{-9.2} & \textbf{115} & \textbf{-7011.28} & \textbf{380.97} & \textbf{613.26} & 474 & \textbf{154343.4}\\

        \bottomrule
    \end{tabular}
    }
    \caption{Training performance of CIER. The goal of this task is to optimize using CIER in different environments based on the DDPG and TD3. We proved the performance of CIER through two sets of experiments. The first set of experiments proved the performance of CIER itself through the basic models and the optimized models, which are represented by -CIER. The baseline models all employ a random experience sampling algorithm. The second set of experiments verified the scalability of CIER and its ability to enhance other experience playback algorithms by comparing PER and CIER approach expanded based on PER, which is represented by -CIPER.}
    
    \label{tab:result}
\end{table*}

\paragraph{TSCF can enhance the effectiveness of experience replay.}

Based on the performance across all environments as presented in Table \ref{tab:result}, our proposed CIER approach consistently outperforms the baseline with statistically significant improvements in average scores across the five environments. Notably, during the early stages of training, our approach demonstrates enhanced effectiveness in guiding the training process, leading to improved convergence speed. This can be attributed to the ability of our approach to identify TSCFs that contribute to reward enhancement, enabling quicker detection and correction of factors leading to score degradation during training. Consequently, our approach achieves higher BS and faster convergence rates.


In general, it is evident that the CIPER approach outperforms the traditional PER approach in several key performance indicators. This demonstrates the effective enhancement of other experience replay algorithms by CIER, highlighting its widespread applicability and efficiency in DRL applications. These results suggest that incorporating the strategies of the CIER approach into PER can significantly improve learning efficiency, reward acquisition, and long-term performance stability.


\paragraph{Explainability of DRL training process with CIER.}

To illustrate the explainability of our approach, we generated the overtaking scenario depicted in Figure \ref{fig:overtake} using the Carla simulation tool. 
We prepared a dataset consisting of 500 instances, encompassing both successful and unsuccessful overtaking scenarios. To present the results more clearly, we opted not to utilize action sequences $T_A$ but instead selected state sequences $T_S$ as inputs. Similar to Equation \ref{equation:reward}, our reward is defined as: 

\begin{equation}
     2 \|\mathbf{L_{c}} - \mathbf{L_{d}}\|_2 + \text {collision},
\end{equation}
where $\|\mathbf{L_{c}} - \mathbf{L_{d}}\|$ represents the \(L_2\) norm between the current position of $T_S$ and the destination of the task, and the collision penalty is set to -1. 


\begin{figure}[h]
	\centering
 \begin{minipage}[]{0.26\textwidth}
     \begin{subfigure}[b]{\textwidth}
     \includegraphics[trim= 1cm 0.5cm 1.5cm 1cm, clip, width=\textwidth]{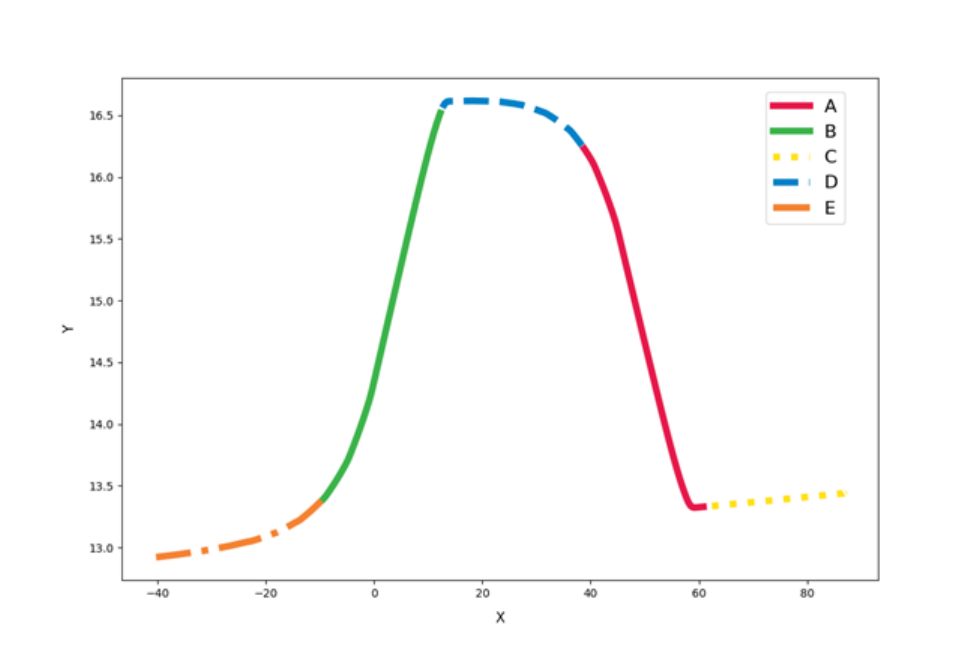}
     \caption{}
     \label{subfig:success}
     \end{subfigure}
      
    \begin{subfigure}[b]{\textwidth}
    \includegraphics[trim= 1cm 0.5cm 1.5cm 1cm, clip, width=\textwidth]{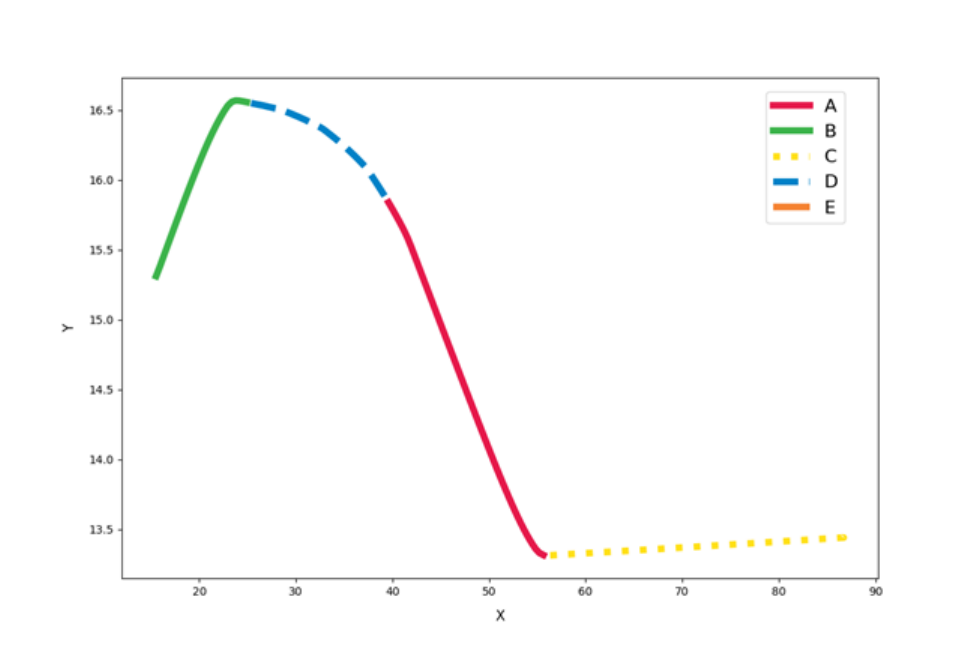}
    \caption{}
    \label{subfig:crash}
    \end{subfigure}
\end{minipage}
\hspace{0.02\textwidth}
\begin{minipage}[]{0.16\textwidth}
  \begin{subfigure}[b]{\textwidth}
  \includegraphics[width=\textwidth]{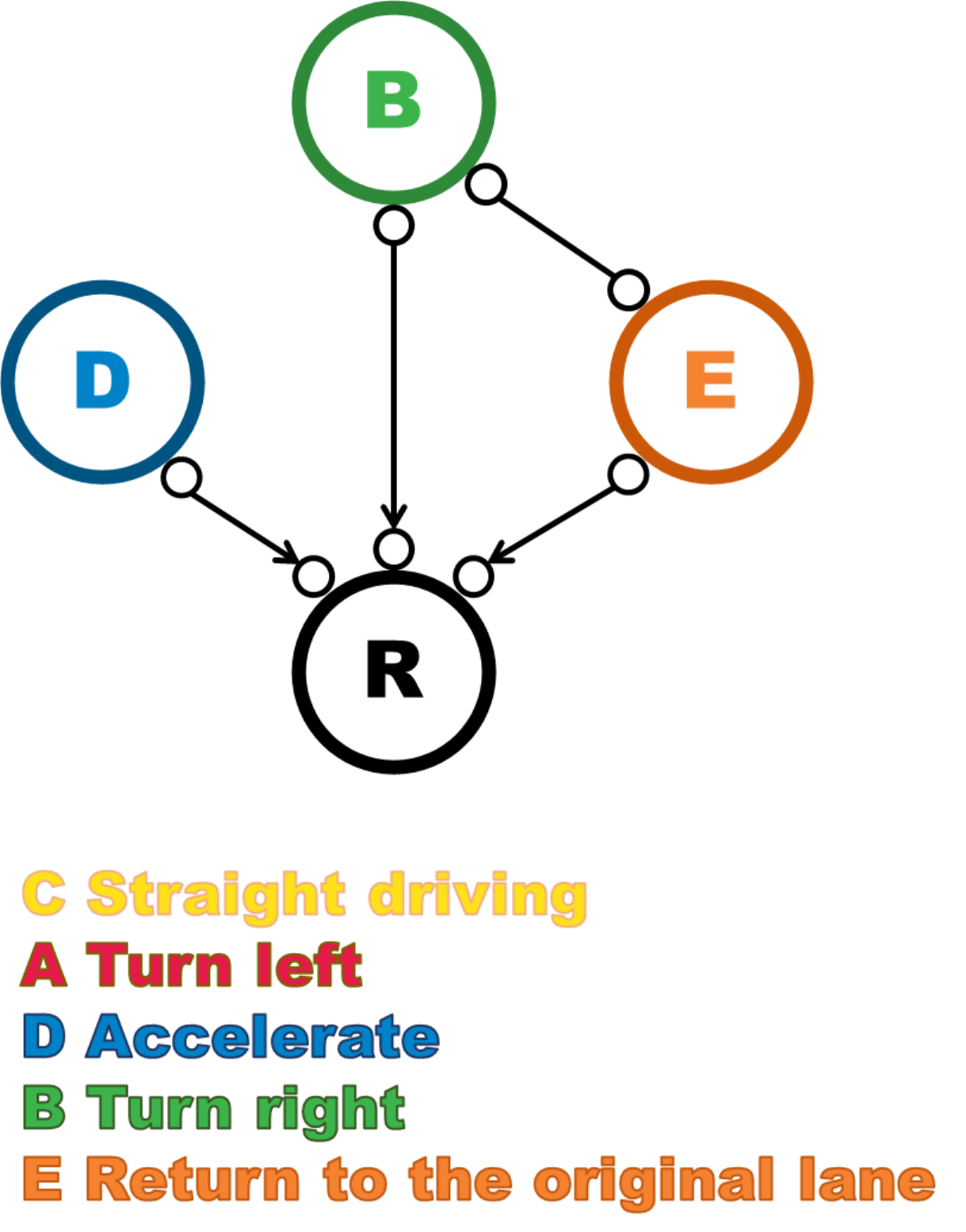}
  \caption{}
  \label{subfig:PAG}
    \end{subfigure}
\end{minipage}
      \caption{TSCFs and PAG for overtaking scenarios. Figure \ref{subfig:success} and \ref{subfig:crash} represent typical trajectory samples of successful overtaking and crash respectively, in which the ego-vehicle travels from the positive direction of the x-axis to the negative direction. Figure \ref{subfig:PAG} is a PAG generated based on GFCI, where treatments correspond to the variables in Figure \ref{subfig:success} and \ref{subfig:crash}, and the explanatory labels in the figures are arranged in chronological order.}
      \label{fig:ticc}
\end{figure}

Through the Time Series Unit and Causal Unit in the CIER approach, we processed the overtaking scenario data, and the resulting outcomes are illustrated in Figure 5. Evidently, in comparison to the hypothesized causal graph in Figure \ref{fig:causal_graph}, except for the straight driving part (\textit{Factor 1}) and the beginning of the turn left part (\textit{Factor 2}) considered irrelevant to the outcome, almost all other factors are almost completely matched. The reason is that for the sake of intuition after trajectory visualization, we fixed the starting position of the vehicle and the early straight action, so the beginning of our scenario is identical in each sample. In the general DRL training process, CIER has the ability to offer similar explanations in each causality-based transition priority correction.


All experiments were conducted on the NVIDIA GeForce RTX 3090 24GB graphics card.

\section{Conclusion}

In this work, we propose a novel algorithm to address the issues of low data efficiency and explainability in the DRL training process. We pay attention to the temporal internal relationships within DRL historical data and represent DRL data using TSCF. Furthermore, we utilize a causal discovery algorithm to calculate the causal relationships between TSCF and DRL learning targets. We construct a causal sum tree to adjust the experience replay strategy. We incorporate the improvement of CIER into the Actor-Critic algorithm across diverse complex DRL environments, demonstrating the effectiveness of our approach. In an extended experiment, we introduce CIER into a PER task, further boosting the performance of DRL training.


\bibliographystyle{named}
\bibliography{ijcai24}

\end{document}